\documentclass[letterpaper, 10 pt, journal, twoside]{IEEEtran}
\IEEEoverridecommandlockouts

\usepackage[nolist]{acronym}

\usepackage[noadjust,compress]{cite}
\usepackage{amsmath,amssymb,amsfonts}
\usepackage{algorithmic}
\usepackage{graphicx}
\usepackage{textcomp}
\usepackage{xcolor}
\usepackage[hypcap=false]{caption}
\usepackage{subcaption}
\usepackage{mathtools}
\usepackage{bm}
\usepackage{booktabs}
\usepackage[inline]{enumitem}
\usepackage{optidef}
\usepackage{hyperref}
\usepackage{import}
\usepackage{tikz}
\usepackage{multirow}
\usepackage{balance}
\usepackage{pgfplots}
\usepackage[permil]{overpic}
\usepackage{pict2e} 
\usepackage{listings}
\usepackage{subcaption}
\pgfplotsset{compat=1.16}

\begin{document}

\newcommand{\equationref}[1]{\hyperref[#1]{Eq.~\ref*{#1}}}
\newcommand{\figref}[1]{\hyperref[#1]{Fig.~\ref*{#1}}}
\newcommand{\tabref}[1]{\hyperref[#1]{Table~\ref*{#1}}}
\newcommand{\secref}[1]{\hyperref[#1]{Section~\ref*{#1}}}
\newcommand{\algoref}[1]{\hyperref[#1]{Algorithm~\ref*{#1}}}
\newcommand{\figsref}[2]{Figures~\ref{#1}-\ref{#2}}
\newcommand{\subfigref}[1]{(\subref{#1})}
\newcommand{\appref}[1]{\hyperref[#1]{Appendix~\ref*{#1}}}

\newcommand{\norm}[1]{\left\lVert#1\right\rVert}
\newcommand{\tbs}[1]{\renewcommand{\tabcolsep}{#1pt}}
\newcommand{\abs}[1]{\left\lvert#1\right\rvert}

\newcommand{\matr}[1]{\mathbf{#1}}
\newcommand*{\prob}{\mathsf{P}}
\newcommand{\de}[1]{\operatorname{d}\!#1}
\newcommand{\etal}[1]{#1 et al.}

\def\bestcolor{(best viewed in color)}
\def\sota{state-of-the-art}
\def\nat{\multicolumn{1}{c}{--}}
\def\baseline{\ac{rrt}-\ac{mc}}
\def\mujoco{MuJoCo}
\def\panda{Frank Panda}

\newcommand{\trsp}{^\mathsf{T}}

\newcommand{\supO}[1]{\prescript{\text{O}}{}{#1}}
\newcommand{\supS}[1]{\prescript{\text{S}}{}{#1}}
\newcommand{\supPi}[1]{\prescript{\text{P}_k}{}{#1}}
\newcommand{\subi}{_{\text{k}}}
\newcommand{\subp}{_{\text{p}}}
\newcommand{\subc}{_{\text{c}}}
\newcommand{\subo}{_{\text{o}}}
\newcommand{\subpi}{_{\text{p}_k}}
\newcommand{\subs}{_{\text{s}}}

\newcommand{\hatbmw}{\hat{\matr{w}}}
\newcommand{\hatbmf}{\hat{\matr{f}}}
\newcommand{\adj}[2]{\text{Ad}_{\matr{T}_{#1 #2}}}
\newacro{mip}[MIP]{Mixed Integer Program}
\newacro{minlp}[MINLP]{Mixed Integer Nonlinear Program}
\newacro{kl}[KL]{Kullback-Leibler}
\newacro{mc}[MC]{Motion Cone}
\newacro{nlp}[NLP]{Nonlinear Program}
\newacro{rrt}[RRT]{Rapidly Exploring Random Tree}
\newacro{slsqp}[SLSQP]{Sequential Least SQuares Programming optimizer}
\newacro{rl}[RL]{Reinforcement Learning}
\newacro{bc}[BC]{Behavior Cloning}
\newacro{mcts}[MCTS]{Monte Carlo Tree Search}

\makeatletter
\let\@oldmaketitle\@maketitle 
\renewcommand{\@maketitle}{\@oldmaketitle 
  \setcounter{figure}{0}
  \vspace{1em}
  \centering   
  \input{Figures/Tikz/pull_figure.tex}
  \vspace{-1em}
  \captionof{figure}{A real-world prehensile pushing example. The trajectory the robot followed is based on our method. As indicated by the arrows, the robot only translates in the three leftmost images, while it translates and rotates in the three rightmost images.}
  \label{fig:pull_figure}
}
\makeatother

\title{Pushing Everything Everywhere All At Once: Probabilistic Prehensile Pushing}
\author{Patrizio Perugini$^{1*}$, Jens Lundell$^{2}$, Katharina Friedl$^{2}$, and Danica Kragic$^{2}$
\thanks{This work was primarily supported by ERC AdV grant BIRD (884807), Knut and Alice Wallenberg Foundation, and the Swedish Research Council. The work was also partially supported by Flanders Make, the research center of manufacturing industry in Belgium \url{https://www.flandersmake.be/en}.}
\thanks{$^{1}$Flanders Make, Belgium. {\tt\footnotesize Patrizio.perugini@flandersmake.be}. $^{*}$Work done while at KTH.}
\thanks{$^{2}$The division of Robotics, Perception, and Learning at KTH, Stockholm, Sweden. 
        {\tt\small \{jelundel,kfriedl,dani\}@kth.se}.}
}

\maketitle

\begin{abstract}
We address prehensile pushing, the problem of manipulating a grasped object by pushing against the environment. Our solution is an efficient nonlinear trajectory optimization problem relaxed from an exact mixed integer non-linear trajectory optimization formulation. The critical insight is recasting the external pushers (environment) as a discrete probability distribution instead of binary variables and minimizing the entropy of the distribution. The probabilistic reformulation allows all pushers to be used simultaneously, but at the optimum, the probability mass concentrates onto one due to the entropy minimization. We numerically compare our method against a state-of-the-art sampling-based baseline on a prehensile pushing task. The results demonstrate that our method finds trajectories 8 times faster and at a 20 times lower cost than the baseline. Finally, we demonstrate that a simulated and real \panda{} robot can successfully manipulate different objects following the trajectories proposed by our method. Supplementary materials are available at \url{https://probabilistic-prehensile-pushing.github.io/}.
\end{abstract}


\section{Introduction}

\IEEEPARstart{P}{rehensile} pushing, exemplified in \figref{fig:pull_figure}, is the task of manipulating grasped objects by pushing them against external \emph{pushers} (the environment) \cite{chavan2015prehensile}. Finding pushing strategies is challenging because of the non-continuous, non-linear dynamics arising from contacts breaking and forming \cite{yuan2018, yuan2019end, pang2023global, jin2024task}. The non-continuity, in particular, discourages gradient-based solutions because the optimization problem turns into an NP-hard \ac{minlp}, which is slow to solve \cite{burer2012non}. As a result, many prior works have addressed the problem using gradient-free sampling-based methods \cite{chavan2020planar,chavan2020sampling, cheng2023enhancing}. Still, sampling-based methods are slow at finding solutions and include many hyper-parameters that must be tuned. 

In this work, we formulate prehensile pushing as a mixed-integer non-linear trajectory optimization problem where pushers are binary integer variables and \acp{mc} \cite{chavan2020planar} are used to calculate admissible prehensile pushes given frictional contact. Furthermore, we propose an efficient continuous non-linear relaxation of the mixed-integer formulation, where the critical insight is recasting the integer variables as a discrete probability distribution and then minimizing the entropy of this distribution. Effectively, the relaxation allows all pushers to be used simultaneously, while the entropy term encourages the optimal solution to converge to one \emph{distinct} pusher. The continuous relaxation is faster to solve and, for specific instances of our problem, will have the same global minimum as the original \ac{minlp}. 

We compare our method numerically against the \sota{} \ac{mc}-based \ac{rrt} planner from \etal{Chavan} \cite{chavan2020planar} on planar non-horizontal prehensile pushing tasks. The results demonstrate that our method finds, on average, a feasible trajectory 8 times faster and at a 20 times lower cost than the baseline. Finally, we validate our method on a simulated and real-world (\figref{fig:pull_figure}) prehensile pushing task. In summary, we contribute:
\begin{itemize}
    \item An exact \ac{minlp} trajectory optimization formulation for prehensile pushing presented in \secref{subsec:MINLP}.
    \item The continuous \ac{nlp} relaxation described in \secref{subsec:relaxation}.
    \item Auxiliary optimization objectives and optimization variables that reduce the pusher switching and speed up the program, presented in \secref{subsec:mv}.
    \item A comprehensive experimental evaluation demonstrating our method's efficacy and real-world applicability, presented in \secref{sec:experiments}.
\end{itemize}

We start by summarizing the related work, followed by the theoretical framework and problem statement.

\section{Related Work}\label{sec:related_work}

Extrinsic dexterity \cite{dafle2014extrinsic} may include pushing against external contact, harnessing gravity, and executing dynamic arm motions. Solutions for extrinsic dexterity are mainly learning-based, sampling-based, or optimization-based, so we review them separately.

\subsection{Learning-Based}
The prevailing learning methods for extrinsic dexterity are \ac{rl} \cite{zhou2023learning,zhou2023hacman,kim2023pre,zhang2023learning,yang2023learning} and \ac{bc} \cite{aceituno2022differentiable}. The \ac{rl}-based methods have mainly focused on object pivoting by either learning direct actions \cite{zhou2023learning,zhou2023hacman,zhang2023learning} or combining predefined manipulation primitives to pivot an object \cite{yang2023learning}. \ac{bc} has been used to learn non-prehensile manipulation from videos \cite{aceituno2022differentiable}. Recently, some works have theoretically shown that \ac{rl} is suitable for learning dexterous manipulation as it stochastically approximates the non-smooth contact dynamics \cite{suh2022differentiable, suh2022bundled}. Thus, the main benefit of learning is that the non-trivial dynamics are implicitly learned rather than explicitly modeled. However, the downside of \ac{rl} is that training is sample-inefficient, and learning often requires fine-tuned rewards, while \ac{bc} requires high-quality demonstrations.

\subsection{Sampling-Based}
If the contact dynamic is explicitly known, we can use it to plan dexterous manipulation through sampling-based or optimization-based methods. Many works  
\cite{chavan2020planar,chavan2020sampling, cheng2023enhancing, pang2023global} have proposed different sampling-based methods as these can better handle the non-continuous contact dynamics than optimization-based methods. Out of these works,  \cite{chavan2020planar,chavan2020sampling,pang2023global} uses different \ac{rrt}-type samplers to find a set of pushes that would bring an object from a start to a goal pose. In contrast, the authors of \cite{cheng2023enhancing} proposed a fast hierarchical planning framework that combines \ac{mcts}, \ac{rrt}, and contact dynamics. 

The sampling-based method most similar to our work is \cite{chavan2020planar}, which uses \acp{mc} in the propagation step of a \ac{rrt}. Our solution also builds upon the \ac{mc} framework but integrates it into a gradient-based trajectory optimization formulation. The primary benefit of our solution over \cite{chavan2020planar} is computational speed.  

\subsection{Optimization-Based}

The main challenge for optimization-based methods is the non-smooth contact dynamic that creates discontinuities in the optimization landscape. Some methods \cite{onol2019contact, mordatch2012contact, mordatch2012discovery, tassa2012synthesis, tassa2014control, chatzinikolaidis2021trajectory} overcome the problem by smoothly approximating the contact dynamics by allowing contacts to act at a distance. Another more physically realistic solution is to model the planning-through-contact problem using complementarity constraints that ensure forces are only non-zero when there is contact \cite{sleiman2019contact,posa2014direct,jin2024task}.

The abovementioned methods solve the trajectory to local optimality, which is significantly easier than to solve for global optimality. There exist some approaches for finding globally optimal contact-rich trajectories \cite{graesdal2024towards,marcucci2017approximate, marcucci2019mixed,hogan2020feedback,hogan2020reactive,aceituno2020global}. All these approaches, except \cite{graesdal2024towards}, formulate the optimization problem as a \ac{mip} that jointly optimizes over discrete contact modes and continuous robotic motions. In comparison, \etal{Graesdal} \cite{graesdal2024towards} frames the problem as finding the shortest path in a graph of convex sets. 
We formulate trajectory optimization as a \ac{minlp}, which we relax into a computationally cheaper \ac{nlp}.

The main difference between our and other optimization-based methods is that we optimize over object twists instead of forces or accelerations. This is possible because the \acp{mc} provides direct dynamic feasibility bounds in the motion space.
\section{Preliminaries}
\label{sec:mc}

Here, we describe \acp{mc}, later used as velocity constraints in our trajectory optimization formulation. The goal is to determine the possible object motions for stable pushing of an object while sliding on a support, as previously described in ~\cite{lynchmason1996, chavan2020planar}. We do not contribute to the \ac{mc} framework but present it here for completeness. 

Consider a 2D rigid object $\matr{O}$ with a body-fixed coordinate frame $O$ whose origin is at the center of mass. Let ${\supO{\matr{w}\subc} = \left[\supO{\matr{f}\subc}\trsp, \supO{\matr{\tau}\subc}\trsp\right]\trsp\in \mathbb{R}^3}$ denote the wrench applied by a contact $c$ and represented in the object frame $O$, as the stacked vector of contact force $\supO{\matr{f}\subc}$ and torque $\supO{\matr{\tau}\subc}$. The object is in frictional contact with a support $s$, with the origin of the support contact frame $S$ located at the center of pressure. In this work, the support is the gripper. A pusher $p$ that interacts with the object is represented by $k=1,\ldots,K$ frictional point contacts $p_k$. The origin of each local friction contact frame $P_k$ lies at the respective contact point. Unless otherwise specified, we denote quantities in the object frame $O$. 

In the present planar scenario, pushing and sliding forces act upon the object within the same plane. Under the quasistatic assumption, i.e., sufficiently slow velocities, inertial effects can be neglected. Then, force balance holds:
\begin{equation}
    \matr{w}\subp + \matr{w}\subs + m\matr{g} = \mathbf{0},
    \label{eq:forcebalance}
\end{equation} with $\matr{w}\subp$ as pusher contact wrench, $\matr{w}\subs$ as support wrench, $m$ as mass of the object, and $\matr{g}$ as gravitational force.

Following \cite{lynchmason1996}, a push is considered stable if contact between the pusher and the object is maintained throughout the motion. Considering friction adhering to Coulomb's law of friction, pushing forces that result in sticking motion at a contact $p\subi$ are constrained to lie within a friction cone $FC\subpi$. The contact normal force $\supPi{\matr{n}\subpi}$ defines the cone's axis of rotation, while the constant friction coefficient $\mu\subp$ determines the opening angle $\beta = \arctan(\mu\subp)$. $\mu\subp$ is assumed the same in static and kinetic friction cases. 
These local effects can be expressed via generalized friction cones in the object's wrench space \cite{erdmann1993friction}. For a pusher $p$, the generalized friction cone is defined as the convex hull:
\begin{equation}
    \matr{W}\subp = \left\{\hatbmw\subpi = \adj{\text{P}_k}{O}\trsp\supPi{\hatbmf\subpi} \;\vert\; \supPi{\hatbmf\subpi}\in FC\subpi, \; \forall k\right\},
    \label{eq:generalized_friction_cone}
\end{equation} where the unit wrench applied by a pusher contact $\hatbmw\subpi$ corresponds to the effect of a unit force $\hatbmf\subpi$ on the object. The adjoint transformation $\adj{\text{P}_k}{O}\trsp$ maps the local pusher contact forces to wrenches in the object frame. 

The maximum set of wrenches on the object by the support contact $\supS{\matr{w}\subs}$ that can be statically transmitted is strictly bounded by an ellipsoidal limit surface. A contact wrench intersecting the limit surface results in a quasistatic sliding motion. The unit support wrench, $\supS{\hatbmw\subs} =\:\supS{\matr{w}\subs}(\mu\subs F_N)^{-1}$, is proportional to the friction coefficient $\mu\subs$ that arises between the object and the support, and magnitude of support normal force $F_N$. It defines the ellipsoidal limit surface as:
\begin{equation}
    \supS{\hatbmw\subs}\trsp\matr{A}\supS{\hatbmw\subs} = 1,
    \label{eq:ellipsoidal_LS}
\end{equation} with $\matr{A} = \text{diag}(\left[1, 1, (re)^{-2}\right])$. Here, $r$ is the radius of the support contact, and $e\in\left[0, 1\right]$ is an integration constant specifying the pressure distribution at the support contact interface.

Combining equations \eqref{eq:forcebalance} and \eqref{eq:generalized_friction_cone}, the force balance condition for a stable push on a sliding object is obtained as:
\begin{equation}
    a\hatbmw\subp + (\mu\subs F_N)\:\adj{S}{O}\trsp \supS{\hatbmw\subs} + m \matr{g} = 0,
    \label{eq:forcebalance_with_generalized_frictioncone}
\end{equation} where $\hatbmw\subp \in \matr{W}\subp,\; a\in\mathbb{R}^+$. The unit pusher wrench inside the generalized friction cone $\hatbmw\subp \in \matr{W}\subp$ is scaled by a magnitude $a$. Note that the magnitude of the support normal force $F_N$ and the design parameters $\mu_s$ and $e$ influence the unique solution of $a$.

\begin{figure}[t]
    \centering
\begin{overpic}[width=0.5\textwidth]{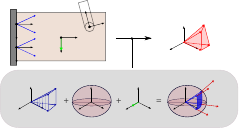}%
\put(25,470) {\scriptsize P}%
\put(380,500) {\scriptsize G}%
\put(380,300) {\scriptsize \textbf{O}}%
\put(80,400) {\scriptsize $p_{0}$}%
\put(80,300) {\scriptsize $p_{1}$}%
\put(235,385){\scriptsize $O$}%
\put(390,380){\scriptsize $S$}%
\put(250,250){\scriptsize (a)}%
\put(130,10){\scriptsize (b)}%
\put(370,10){\scriptsize (c)}%
\put(570,10){\scriptsize (d)}%
\put(770,10){\scriptsize (e)}%
\put(770,250){\scriptsize (f)}%
\put(558,160){\scriptsize  y}%
\put(540,65){\scriptsize z}%
\put(620,90){\scriptsize x}%
\put(310,390){\scriptsize x}%
\put(235,320){\scriptsize z}%

\end{overpic}
   \caption{Visual description for calculating \acp{mc}. (a) Shows the pusher (P) pushing the rectangular object (\textbf{O}), which slides inside the gripper (G). Together, the convex hull of the generalized friction cones (b), the limit surface (c), and the gravity acting on the object (d) result in the motion resolution (e). The final polyhedral approximation of the motion cone is shown in (f). The axes labels in (b), (c), (e), and (f) follow the convention in (d), where forces act along the x- and z-axis and torque around the y-axis \bestcolor{}.}
   \label{fig:preliminaries}
\end{figure}

Solving equations \eqref{eq:ellipsoidal_LS} and \eqref{eq:forcebalance_with_generalized_frictioncone} together yields a set $\supS{\matr{W}\subs}$, which effectively constrains the admissible support contact wrenches $\supS{\hatbmw\subs} \in \supS{\matr{W}\subs}$ during stable pushing of a sliding object. For a uniform and finite support friction distribution, the resulting sliding velocity from such a frictional contact wrench is normal to its intersection point with the limit surface \cite{goyal1991LS}. Hence, the direction of the velocity can be obtained via differentiation of \eqref{eq:ellipsoidal_LS}. Let the object motion w.r.t. the support frame be denoted by the twist $\supS{\bm{\xi}\subo} = \left[\supS{\matr{v}\subo}\trsp, \supS{\matr{\omega}\subo}\trsp\right]\trsp\in \mathbb{R}^3$, with linear velocity $\supS{\matr{v}\subo}$ and angular velocity $\supS{\matr{\omega}\subo}$. Then, the set of possible object motions resulting from stable pushes is defined as the motion cone:
\begin{equation}
    \supS{\matr{V}\subo} = \left\{\supS{\bm{\xi}\subo} = b \matr{A} \supS{\hatbmw\subs} \;\vert\; \supS{\hatbmw\subs} \in \supS{\matr{W}\subs}, b \in \mathbb{R}^+\right\}, 
    \label{eq:motioncone}
\end{equation} with velocity scaling factor $b$. Note that the wrench is independent of the speed of motion \cite{kao2016contact}. An illustration of these preliminaries is provided in Figure~\ref{fig:preliminaries}.

To compute the motion cone, we follow \cite{chavan2020planar} and first sweep over the boundaries of the pusher cone, obtain the boundaries of support and motion cone, and approximate the surfaces connecting supports linearly to end up with a polyhedral approximation of this motion cone. In our following formulation of the trajectory optimization problem, we rely on the computation of motion cones as the set of admissible object twists that will result in stable pushes and apply the motion cone as a constraint.

\section{Problem Statement}\label{sec:problem_statement}

Using the same notation as in the previous section, the problem we address is manipulating $\matr{O}$ from a start pose $\matr{P}_0 \in \text{SE}(2)$ to a goal pose $\matr{P}_g \in \text{SE}(2)$ by sequentially pushing against external pushers $p^{0},\dots,p^{M} \in \{0,1\}$. The push action is a twist $\bm{\xi}$. The goal is reached when $d(\matr{P}_N,\matr{P}_g) \leq \epsilon$, where $d: \text{SE}(2) \times \text{SE}(2) \rightarrow \mathbb{R}_{\geq 0}
$ is a distance function, $\matr{P}_N$ is the final pose and $\epsilon$ is a user-specified distance threshold. We define the distance function as $d(\matr{P}_N,\matr{P}_g)=\lambda_s\norm{\bm{\kappa}_N-\bm{\kappa}_g}+\lambda_\theta\abs{\theta_N-\theta_g}$, where $\bm{\kappa}\in\mathbb{R}^2$ is a position vector, $\theta$ is a rotation around the global coordinate system's y-axis, and $\lambda_s$ and $\lambda_\theta$ are user-specified weights.

We represent $\matr{O}=\left\{\bm{\nu}_0, \bm{\nu}_1, \dots, \bm{\nu}_I\right\}$ with a set of vertices $\bm{\nu}_i \in \mathbb{R}^2$, where $\bm{\nu}_i$ is connected to $\bm{\nu}_{i+1}$ and $\bm{\nu}_I$ is connected to $\bm{\nu}_0$.
We predefine the number of pushing sequences $N$ and the maximum execution time $T$, which are used to calculate the time-step for each push as $\delta=\frac{T}{N-1}$. Given these specifications, the problem becomes finding a sequence of pushers $p^{0},\dots,p^{M}$ and twists $\bm{\xi}_0,\dots,\bm{\xi}_{N-1}$ that minimizes $d(\matr{P}_N,\matr{P}_g)$.
\section{Prehensile Pushing as Trajectory Optimization}
\label{sec:method}
We now present the \ac{minlp} formulation for prehensile pushing and then the computationally more tractable continuous \ac{nlp} relaxation. Finally, we discuss additional cost and optimization variables used to speed up the optimization and find trajectories that are easier to execute on physical hardware.

\subsection{Mixed Integer Nonlinear Programming Formulation}\label{subsec:MINLP}

We define the trajectory optimization problem as minimizing the distance between $\matr{P}_N$ and $\matr{P}_g$ while keeping the twists inside the motion cone and the pusher on the object. Mathematically, this is formulated as:

\begin{mini!}|s|%
{\substack{\strut \bm{\xi}^m_n,~\matr{P}_n,~p^m_n \\ m\in\{0,\ldots,M\}\\ n\in\{0,\ldots,N-1\}}}
 { \|\matr{P}_{N} - \matr{P}_{g}\|+\lambda_{p} \sum_{n=0}^{N-1} \|\matr{P}_{n+1} - \matr{P}_{n}\|,\label{eq:minlp}}%
{\label{optipb}}%
{}
\addConstraint{\bm{\xi}_{n}^{m} \in \supS{\matr{V}_{o,n}^{m}}\label{eq:nonlinear_const}}
\addConstraint{ \matr{P}_n \in \matr{O}\label{eq:state_const}}
\addConstraint{\matr{P}_{n+1}= \matr{P}_n+\delta\sum_{m = 0}^{M} \bm{\xi}_n^{m}p_{n}^{m}\label{eq:dynamics_const}}
\addConstraint{\sum_{m = 0}^{M} p_{n}^{m} = 1\label{eq:sum_const}}
\addConstraint{p_{n}^{m} \in \{0,1\}, \label{eq:integer_const}}
\end{mini!}
where $n$ describes the discretization point of the trajectory, $\bm{\xi}^m_n$ is the m-th twist at the n-th segment, $\supS{\matr{V}_{o,n}^{m}}$ is the m-th motion cone at the n-th segment, $p_{n}^{m}$ is the m-th pusher at the n-th segment, and $\lambda_{p}$ is a user-specified weighting term. In the above optimization problem, \equationref{eq:nonlinear_const} ensures the twist is inside the \ac{mc}, \equationref{eq:state_const} ensures the gripper is on the object, \equationref{eq:dynamics_const} ensures the dynamics hold, and \equationref{eq:sum_const} and \equationref{eq:integer_const} combined ensure that one and only one pusher is active. 

If $\matr{O}$ is convex, we can express \equationref{eq:state_const} as the linear constraint:
\begin{align}
\label{eq:linear_state_const}
\matr{A}
\begin{pmatrix}
x_n\\y_n
\end{pmatrix}
\leq \matr{b},
\end{align}
where $\matr{A} \in \mathbb{R}^{I\times 2}$ and $\matr{b} \in \mathbb{R}^{I}$ are derived from the vertices of $\matr{O}$ and $x_n$ and $y_n$ are the 2D position of $\matr{P}_n$. Non-convex $\matr{O}$ cannot be expressed in the same linear form. Instead, for such objects, we first divide the object into a set of convex regions $\{ \matr{U}_0, \dots, \matr{U}_L | \bigcup_{l=0}^ L \matr{U}_l=\matr{O}\}$ where each $\matr{U}_l$ is associated with a linear state-constraint $\matr{A}_l
\begin{pmatrix}
x_n\\y_n
\end{pmatrix}-\matr{b}_l
\leq 0$ and a binary variable $\psi_l \in \{0,1\}$ that is 1 iff $\matr{P}_n \in \matr{U}_l$. Then, we can form the state constraint as the integer constraint:
\begin{align}
\label{eq:integer_state_const}
\begin{split}
\sum_{l = 0}^{L} \psi_{l} &= 1,\\
\psi_{l}\left[\matr{A}_l
\begin{pmatrix}
x_n\\y_n
\end{pmatrix} -\matr{b}_l\right]
&\leq 0~\forall l \in \{0, \dots, L\}.
\end{split}
\end{align}

The optimization problem in \equationref{eq:minlp} is a \ac{minlp} due to the integer constraints \eqref{eq:integer_const} and \eqref{eq:integer_state_const}, and the non-linear constraint \eqref{eq:nonlinear_const}. Unfortunately, \acp{minlp} are NP-hard \cite{burer2012non}, meaning they are computationally expensive to solve. Thus, to make the problem computationally more tractable, we propose to continuously relax \equationref{eq:minlp} into a \ac{nlp}.

\subsection{Continuous Relaxation}\label{subsec:relaxation}

We continuously relax the integer variables $p_n^m$ in \equationref{eq:integer_const} by treating them as probabilities. This reformulation has one obvious limitation: it allows for solutions where the robot simultaneously uses all pushers, which is physically impossible. Therefore, the probability mass should concentrate on one distinct $p_n^m$ at the optimum. To encourage this, we add the entropy of the discrete probability distribution as an additional cost term. Mathematically, the continuous relaxation of \equationref{eq:integer_const} becomes:
\begin{mini}|s|
{\substack{\strut \bm{\xi}^m_n,~\matr{P}_n,~p^m_n \\  m\in\{0,\ldots,M\}\\ n\in\{0,\ldots,N-1\}}}
{ \|\matr{P}_{N} - \matr{P}_{g}\| +\lambda_{p} \sum_{n=0}^{N-1} \|\matr{P}_{n+1} - \matr{P}_{n}\|}{}{}
\breakObjective {+ \lambda_{e} \sum_{n=0}^{N-1}\sum_{m=0}^{M} p_{n}^{m}\log(p_{n}^{m})}%
{\label{eq:relaxation}}
\addConstraint{\eqref{eq:nonlinear_const}-\eqref{eq:sum_const}}{}
\addConstraint{p_{n}^{m} \geq 0,}{}
\end{mini}
where $\lambda_{e}$ is a user-specified weighing term. Ideally, the entropy term in \equationref{eq:relaxation} concentrates the probability mass onto one pusher. Thus, once we execute the actual trajectory, we sample the most probable pushers as:
\begin{equation}
p_n^* = \operatorname*{arg\,max}_{m \in \{0, ..., M\}} p_n^m,~\forall n\in\{0,\dots,N\}.
\label{eq:rounding}
\end{equation}
Despite the rounding in \equationref{eq:rounding} the constraints \eqref{eq:nonlinear_const}-\eqref{eq:sum_const} remain feasible because 
\begin{enumerate*}
   \item $\bm{\xi}_n^m \in \supS{\matr{V}_{o,n}^{m}}$ is enforced during optimization, satisfying constraint \eqref{eq:nonlinear_const},
   \item $\matr{P}_n \in \matr{O}$ is independent of the pusher selection, satisfying constraint \eqref{eq:state_const},
   \item the dynamics remain valid as we only execute motions corresponding to the selected pusher, satisfying constraint \eqref{eq:dynamics_const}, and
   \item exactly one pusher is selected at each timestep through the $\arg\max$ operation, satisfying constraint \eqref{eq:sum_const}.
\end{enumerate*}

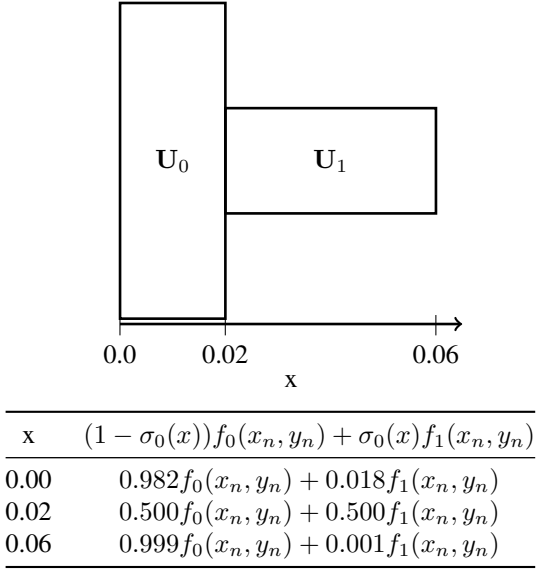
\begin{figure}[tb]
    \centering
    \begin{tikzpicture}[baseline]
    \begin{scope}[scale=70]
        \draw[->, line width=1pt] (0,-0.001) -- (0.065,-0.001);
        \node[below] at (0.065/2,-0.009) { x};
    
        \foreach \x in {0.0,0.02,0.06}
            \draw (\x,0.02pt) -- (\x,-0.09pt) node[below] {\x};
    
        \draw[line width=1pt] (0.00002,0.00002) rectangle (0.02002,0.06002);
        \draw[line width=1pt] (0.02002,0.02002) rectangle (0.06002,0.04002);
        
        \node[font=\bfseries] at (0.01002,0.03002) {$\matr{U}_0$};
        \node[font=\bfseries] at (0.04002,0.03002) {$\matr{U}_1$};
    \end{scope}
    
    \node[anchor=north] at (2,-1.1) {
        \begin{tabular}{@{}cc@{}}
            \toprule
            x & $(1-\sigma_0(x))f_0(x_n,y_n)+\sigma_0(x)f_1(x_n,y_n)$ \\
            \midrule
            0.00 & $0.982f_0(x_n,y_n)+0.018f_1(x_n,y_n)$ \\
            0.02 & $0.500f_0(x_n,y_n)+0.500f_1(x_n,y_n)$ \\
            0.06 & $0.999f_0(x_n,y_n)+0.001f_1(x_n,y_n)$ \\
            \bottomrule
        \end{tabular}
    };
\end{tikzpicture}
    \caption{The smoothed non-convex state constraint for the T-shaped object. Here, $\alpha=200$ and $\hat{x}_0=0.02$.}
    \label{fig:example_non_convex_state_constraint}
\end{figure}

We propose to smooth the boundaries between the convex regions to relax the integer non-convex state-constraint \eqref{eq:integer_state_const}. Specifically, for each neighboring convex regions $\matr{U}_l$ and $\matr{U}_{l+1}$ we add a Sigmoid function:
\begin{align}
    \sigma_l(x) = \frac{1}{1 + \exp(-\alpha \cdot (x - \hat{x}_l))},
    \label{eq:sigmoid}
\end{align}
where $\hat{x}_l$ is the x-value between the two convex regions and $\alpha$ is a hyperparameter. Using the Sigmoid functions, we can formulate the continuous non-convex state constraint as:
\begin{align}
\label{eq:continuous_non_convex_state_const}
\begin{split}
    (1-\sigma_0(x_n))(1-\sigma_1(x_n))\cdots(1-\sigma_{L-1}(x_n))f_0(x_n,y_n) + \\
    \sigma_0(x_n)(1-\sigma_1(x_n))\cdots(1-\sigma_{L-1}(x_n))f_2(x_n,y_n) + \dots +\\ \sigma_0(x_n)\sigma_1(x_n)\sigma_{L-1}(x_n)f_{L-1}(x_n,y_n) \leq 0,
\end{split}
\end{align}
where $f_l(x_n, y_n)=\matr{A}_l
\begin{pmatrix}
x_n\\y_n
\end{pmatrix}- \matr{b}_l$. By tuning $\alpha$ in \equationref{eq:sigmoid}, we can make the boundary infinitely steep, and at $\alpha \to \infty$, we recover the integer constraint in \equationref{eq:integer_state_const}. \figref{fig:example_non_convex_state_constraint} shows an example of the smoothed non-convex state constraint with only one $\sigma$. 

The relaxed problem can now be solved using gradient-based optimization in polynomial time. The relaxation in \equationref{eq:relaxation} is also conservative because it has the same global optimum as \equationref{eq:minlp}, assuming such an optimum exists and that $N \to \infty$. This is true because the minimization of the entropy cost results asymptotically in a discrete probability distribution. The initial solution begins around the barycenter of the simplex and evolves towards one of the vertices, where only one mode remains active. Such vertex coincides with the global minimum of the original \ac{minlp}.

\subsection{Additional Improvements}\label{subsec:mv}

Solving \equationref{eq:relaxation} already results in physically plausible trajectories that, once executed, manipulate the object. Still, we can further speed up the computation and ensure that the trajectories are more efficient to execute on a real robot if we
\begin{enumerate*}[label=(\roman*)]
    \item allow for variable execution time $T$, and
    \item minimize pusher switching.
\end{enumerate*}

\textbf{Variable Time}. Instead of prespecifying $T$, we add it as an additional optimization variable. Adding it gives the optimizer more freedom at the cost of an additional optimization variable and constraint $\delta = \frac{T}{N-1}$.

\textbf{Pusher Switch Minimization}. \equationref{eq:relaxation} allows for arbitrarily many pusher switches. However, on a real system where switching pushers is costly, the fewer, the better. We incorporate pusher switch minimization into \equationref{eq:relaxation} by adding the \ac{kl} divergence between consecutive probability distributions as a cost. Mathematically, we add the following objective 
\begin{equation}\label{eq:kl}
\lambda_{kl}\sum_{n=0}^{N-1}D_{KL}(p_{n+1}^m||p_{n}^m),
\end{equation}
where $D_{KL}(p_{n+1}^m||p_{n}^m)= \sum_{m=0}^{M}p_{n+1}^m \log \frac{p_{n+1}^m}{p_{n}^m}$.

Incorporating all the additional objectives and constraints results in the optimization problem:
\begin{mini}|s|
{\substack{\strut \bm{\xi}^m_n,~\matr{P}_n,~p^m_n~T \\  m\in\{0,\ldots,M\}\\  n\in\{0,\ldots,N-1\}}}{ \|\matr{P}_{N} - \matr{P}_{g}\| + \lambda_{p} \sum_{n=0}^{N-1} \|\matr{P}_{n+1} - \matr{P}_{n}\|}{}{}
\breakObjective {+\lambda_{e} \sum_{n=0}^{N-1}\sum_{m=0}^{M} p_{n}^{m}\log(p_{n}^{m})}
\breakObjective {+\lambda_{kl}\sum_{n=0}^{N-2}\sum_{m=0}^{M}D_{KL}(p_{n+1}^m||p_{n}^m)}
{\label{eq:full_relaxation}}{}
\addConstraint{\eqref{eq:nonlinear_const}-\eqref{eq:sum_const}}{}
\addConstraint{p_{n}^{m} \geq 0}{}
\addConstraint{\delta=\frac{T}{N-1},}{}
\end{mini}
where $\lambda_p$, $\lambda_e$, and $\lambda_{kl}$ are user-specified weighing terms for the separate loss functions.
\section{Experiments}\label{sec:experiments}

We experimentally evaluate our trajectory optimization formulation on a 2D non-horizontal prehensile pushing task\footnote{Source code is openly available at \url{https://probabilistic-prehensile-pushing.github.io/}}. To solve the \acp{nlp}, we used the \ac{slsqp} version 1.9.1 from SciPy \cite{virtanen2020scipy}. We used the following hyperparameters in all experiments as they delivered the best performance: $N=3$, $\alpha=200$, $T=2$, $\lambda_p=\lambda_e=\lambda_{kl}=0.1$, $\lambda_s=0.9$, and $\lambda_\theta=0.1$. All computations are done on Ubuntu 20.04.5 LTS using an Intel Core i7-10750H 2.60GHz processor. $\matr{P}_n$ are initialized using linear interpolation between $\matr{P}_0$ and $\matr{P}_N$. The twists $\bm{\xi}^m_n$ are initialized using finite differencing $\bm{\xi}^m_n= \frac{1}{\delta}\log(\matr{P}_{n-1}^{-1}\matr{P}_n)$. Pusher probabilities $p^m_n$ are initialized uniformly ($p^m_n = \frac{1}{M+1}$) to avoid bias towards any particular pusher.

The specific questions we wanted to answer with the experiments were:
\begin{enumerate}
    \item What are the differences between our approach and \baseline{} regarding computational time and solution quality? 
    \item What effects do the variable time and switch minimization have on the computational efficiency and the solution?
    \item How successful is a Franka Panda robot at manipulating objects when following trajectories generated by our method?
\end{enumerate}

\subsection{Numerical Evaluation}

\label{sec:numerical_evaluation}We numerically evaluated all methods on a convex squared object and a non-convex T-shaped and L-shaped object. We fixed the goal state for each object while uniformly sampling 20 different start states. An example of the start and goal states for the T-shape is shown in \figref{fig:random_sampling}. The environment included four different pushers. We benchmarked against our implementation of the \ac{rrt}-based \ac{mc} method from \cite{chavan2020planar}, hereafter referred to as \baseline{}. We set $\epsilon=10^{-4}$. 

\begin{figure}[t]
    \centering
    \begin{subfigure}[b]{0.24\textwidth}
        \centering
        \resizebox{\textwidth}{!}{


\begin{tikzpicture}
\begin{axis}[
    scale only axis,
    xmin=-0.035, xmax=0.035,
    ymin=-0.035, ymax=0.035,
    axis lines=left,
    xlabel={\Large x},
    ylabel={\Large y},
    xtick={-0.02,0,0.02},
    ytick={-0.02,0,0.02},
    tick style={line width=1pt},
    legend columns=1,
    font=\LARGE,
    align=left,
    legend cell align=left,
    x axis line style={line width=2pt},
    y axis line style={line width=2pt},
    legend style={
        at={(0.4,-0.18)},
        anchor=north,
        legend columns=3,
        /tikz/every even column/.append style={column sep=0.5cm}
    },
    clip=false,
    scaled ticks=false,
    xticklabel style={
        /pgf/number format/fixed,
        /pgf/number format/precision=2,
        /pgf/number format/fixed zerofill
    },
    yticklabel style={
        /pgf/number format/fixed,
        /pgf/number format/precision=2,
        /pgf/number format/fixed zerofill
    }
]

\addplot[color=black, only marks, mark=triangle*, mark size=3.5pt] table {Data/start_positions.txt};

\addplot[color=black, only marks, mark=diamond*, mark size=3.5pt] 
    coordinates {(0.02,-0.003)};

\addplot[draw=black, line width=2pt, fill=none] coordinates {(-0.03,-0.03) (-0.01,-0.03) (-0.01,0.03) (-0.03,0.03) (-0.03,-0.03)};
\addplot[draw=black, line width=2pt, fill=none] coordinates {(-0.01,-0.01) (0.03,-0.01) (0.03,0.01) (-0.01,0.01) (-0.01,-0.01)};

\addlegendentry{Uniformly Sampled Start Positions}
\addlegendentry{Goal Positions}

\end{axis}
\end{tikzpicture}

}
        \caption{}
        \label{fig:random_sampling}
    \end{subfigure}
    \hfill
    \begin{subfigure}[b]{0.24\textwidth}
        \centering
        \resizebox{\textwidth}{!}{


\begin{tikzpicture}
\begin{axis}[
    scale only axis,
    xmin=-0.035, xmax=0.035,
    ymin=-0.035, ymax=0.035,
    axis lines=left,
    xlabel={\Large x},
    ylabel={\Large y},
    legend columns=2,
    font=\LARGE,
    align=left,
    xtick={-0.02,0,0.02},
    ytick={-0.02,0,0.02},
    tick style={line width=1pt},
    x axis line style={line width=2pt},
    y axis line style={line width=2pt},
    legend style={
        at={(0.45,-0.18)},
        anchor=north,
        legend columns=3,
        /tikz/every even column/.append style={column sep=0.5cm}
    },
    clip=false,
    scaled ticks=false,
    xticklabel style={
        /pgf/number format/fixed,
        /pgf/number format/precision=2,
        /pgf/number format/fixed zerofill
    },
    yticklabel style={
        /pgf/number format/fixed,
        /pgf/number format/precision=2,
        /pgf/number format/fixed zerofill
    }
]

\addplot[color=black, only marks, mark=triangle*, mark size=3.5pt] 
    coordinates {(-0.02,-0.025)};

\addplot[color=black, only marks, mark=diamond*, mark size=3.5pt] 
    coordinates {(0.02,-0.003)};

\addplot[color=black, mark=*, mark size=2.5pt, line width=1.5pt] 
    coordinates {(-0.018,0.005) (0.01,0.00)};

\addplot[color=black, mark=none, line width=1.5pt] coordinates {(-0.02,-0.025) (-0.018,0.005)};
\addplot[color=black, mark=none, line width=1.5pt] coordinates {(0.01,0.00) (0.02,-0.003)};

\addplot[draw=black, line width=2pt, fill=none] coordinates {(-0.03,-0.03) (-0.01,-0.03) (-0.01,0.03) (-0.03,0.03) (-0.03,-0.03)};
\addplot[draw=black, line width=2pt, fill=none] coordinates {(-0.01,-0.01) (0.03,-0.01) (0.03,0.01) (-0.01,0.01) (-0.01,-0.01)};

\addlegendentry{Start Position}
\addlegendentry{Goal Position}
\addlegendentry{Gripper Positions}

\end{axis}
\end{tikzpicture}

}
        \caption{}
        \label{fig:example_trajectory}
    \end{subfigure}
    \caption{(\subref{fig:random_sampling}) shows the uniformly sampled start and goal positions while (\subref{fig:example_trajectory}) shows an optimized trajectory.}
    \label{fig:combined_figure}
\end{figure}
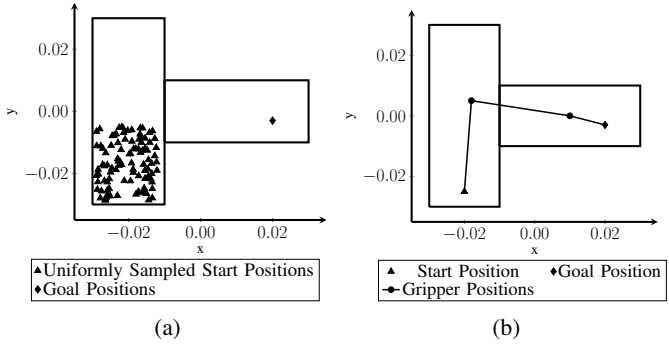

\begin{figure}[t]
    \centering
    \resizebox{0.9\columnwidth}{!}{    \begin{tikzpicture}
\begin{semilogyaxis}[
    scale only axis,
    xlabel={Time (s)},
    ylabel={Goal Distance (mm)},
    axis lines=left,
    xlabel shift={-3pt},
    ylabel shift={-5pt},
    xmin=0.0, xmax=160,
    ymin=1e-5, ymax=1e2,
    x axis line style={line width=1pt},
    y axis line style={line width=1pt},
    legend style={
        at={(0.5,-0.15)},
        anchor=north,
        legend columns=3,
        /tikz/every even column/.append style={column sep=0.5cm}
    },
    clip=false,
    tick label style={font=\footnotesize},
    legend cell align={left},
]

\addplot[black, thick, mark=triangle*, mark options={scale=1.5}] coordinates {
    (15, 1000*1e-2) (43, 1000*1e-3) (87, 1000*1e-4) (160, 1000*1e-5)
};
\addlegendentry{RRT-MC (Square)}

\addplot[black, thick, dashed, mark=triangle*, mark options={scale=1.5, solid}] coordinates {
    (18, 1000*1e-2) (63, 1000*1e-3) (140, 1000*1e-4)
};
\addlegendentry{RRT-MC (T)}

\addplot[black, thick, dotted, mark=pentagon*, mark options={scale=1.5, solid}] coordinates {
    (21, 1000*1e-2) (75, 1000*1e-3) (160, 1000*1e-4)
};
\addlegendentry{RRT-MC (L)}

\addplot[black, thick, mark=diamond*, mark options={scale=1.5}] coordinates {
    (20, 1000*1e-6) (30, 1000*1e-7)
};
\addlegendentry{Ours (Square)}

\addplot[black, thick, dashed, mark=diamond*, mark options={scale=1.5, solid}] coordinates {
    (24, 1000*1e-5) (37, 1000*1e-6)
};
\addlegendentry{Ours (T)}

\addplot[black, thick, dotted, mark=pentagon*, mark options={scale=1.5, solid}] coordinates {
    (26, 100*1e-4) (37, 100*1e-5)
};
\addlegendentry{Ours (L)}
\end{semilogyaxis}
\end{tikzpicture}}
    \caption{The relationship between the goal distance and the computational budget. T stands for the T-shaped and L for the L-shaped object.}
    \label{fig:RRT_vs_traj_opt}
\end{figure}
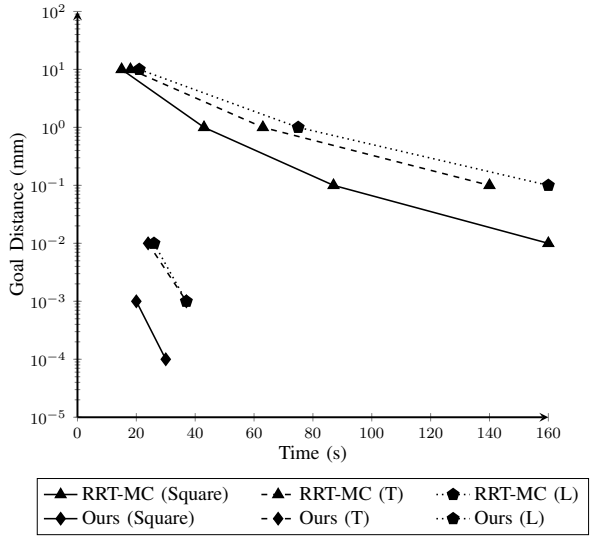

The numerical results are shown in \figref{fig:RRT_vs_traj_opt}. The results demonstrate that, after 20 seconds, our method's goal distance is 1,000 times smaller for the non-convex object and 10,000 times smaller for the convex one than \baseline{}. Even after running \baseline{} for 140 seconds on the convex and 160 seconds on the non-convex object, the distance to the goal is still 100 times higher than the solution our method provides after only 20 seconds. \figref{fig:example_trajectory} shows an example trajectory found using our method. 

We further analyzed how different optimization variables impact the computational time and the goal distance when optimizing \equationref{eq:relaxation}. The results are presented in \tabref{tab:ablation} and demonstrate that the direct transcription formulation, where both $\bm{\xi}$ and $\matr{P}$ are optimized, leads to a ten times lower goal distance than single shooting methods that only optimize $\bm{\xi}$. Adding $T$ as an optimization variable reduces the computational time by 2--4 seconds. 

\begin{figure}[tb]
    \centering
    \resizebox{0.7\columnwidth}{!}{\begin{tikzpicture}
\begin{axis}[
    scale only axis,
    axis lines=left,
    xlabel={\LARGE $\mu$},
    ylabel={\large $Q$ (\%)},
    xmin=0.4, xmax=1,
    ymin=0, ymax=16,
    xtick={0.4,0.5,0.6,0.7,0.8,0.9,1.0},
    ytick={0,5,10,15},
    legend style={
        at={(0.5,-0.20)},
        anchor=north,
        legend columns=1,
        font=\large
    },
    x axis line style={line width=1.5pt},
    y axis line style={line width=1.5pt},
    tick style={line width=1pt},
]

\addplot[
    color=black,
    line width=2pt,
    mark=*
] coordinates {
    (0.4, 15)
    (0.5, 14)
    (0.6, 11)
    (0.7, 7)
    (0.8, 3)
    (0.9, 0.3)
    (0.99, 0.1)
};
\addlegendentry{Switch ratio}

\end{axis}
\end{tikzpicture}}
    \caption{The reduction in switches w/ and w/o the \ac{kl} penalty as a function of the friction coefficient $\mu$.}
    \label{fig:kl}
\end{figure}
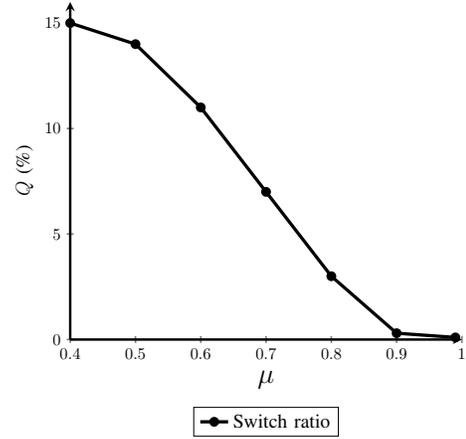

Finally, we investigated if the \ac{kl}-penalty in \equationref{eq:full_relaxation} reduced the number of pusher switches. To quantify the reduction in pusher switches, we propose the metric $Q=100\times\frac{\Delta-\Delta_{kl}}{\Delta}$, where $\Delta_{kl}$ and $\Delta$ are the total number of pusher switches with and without the \ac{kl} penalty, respectively. The Q-metric represents the percentage reduction in pusher switches when using the KL penalty compared to not using it, where a higher value indicates fewer pusher switches. \figref{fig:kl} illustrates the switch reduction as a function of $\mu$. The results demonstrate that with $\mu=0.4$, the \ac{kl} penalty decreases pusher switching with $15\%$. As expected, the percentage of switches approaches zero as the friction coefficient increases because the \ac{mc} becomes wide enough that object manipulation is possible using only one pusher.

\begin{table}[tb]
 \caption{Solution time and objective value for different optimization variables. T stands for the T-shaped and L for the L-shaped object.}
 \begin{center}
 \begin{tabular}{l ccc ccc}
 \toprule
 \multirow{2}{*}{\shortstack[l]{\textbf{Optimization}\\\textbf{Variables}}} & \multicolumn{3}{c}{\textbf{Time (s)}} & \multicolumn{3}{c}{\textbf{Objective Value}} \\
 \cmidrule(lr){2-4} \cmidrule(lr){5-7} 
& \textbf{Square} & \textbf{T} & \textbf{L} & \textbf{Square} & \textbf{T} & \textbf{L} \\
  \midrule
$\bm{\xi},~\matr{p}$ & 23 & 27 & 26 & $e^{-6}$ & $e^{-6}$ & $e^{-6}$ \\
$\bm{\xi},~\matr{p},~T$ & 20 & 23 & 23 & $e^{-6}$ & $e^{-6}$ & $e^{-6}$ \\
$\bm{\xi},~\matr{p},~\matr{P}$ & 34 & 39 & 37 & $e^{-7}$ & $e^{-6}$ & $e^{-6}$ \\
$\bm{\xi},~\matr{p},~\matr{P},~T$ & 31 & 37 & 36 & $e^{-7}$ & $e^{-7}$ & $e^{-6}$ \\
 \bottomrule
 \end{tabular}
 \label{tab:ablation}
 \end{center}
 \end{table}

\subsection{Evaluation in Physics Simulation}

Next, we evaluate \baseline{} and our best method from the numerical experiments, namely \equationref{eq:full_relaxation}, in the physics engine \mujoco{} \cite{todorov2012mujoco}. The simulation environment includes a \panda{} robot, the object to manipulate, and a free-floating fixture that serves as the external pusher\footnote{Videos of the environment are available at \url{https://probabilistic-prehensile-pushing.github.io/}}. The end-effector twists were mapped to joint velocities using differential inverse kinematics. We performed a grid search over \mujoco's hyper-parameters to find a combination that enabled prehensile pushing. The selected hyper-parameters are detailed in \appref{app:technical_details}.

Each method generated five different trajectories for each object type (square, T-shaped, and L-shaped), using consistent pre-selected start and goal poses. The evaluation used two $\epsilon$ thresholds: $\epsilon=5^{-4}$ for high-precision control and $\epsilon=6^{-3}$ for standard operation. With the stricter threshold ($\epsilon=5^{-4}$), our method achieved successful manipulation in all trials (100\% success rate), while \baseline{} failed to reach the target poses (0\% success). When relaxing the threshold to $\epsilon=6^{-3}$, both methods achieved complete success (100\%). These results demonstrate that our method can generate robust trajectories under realistic physics simulation, particularly for high-precision manipulation tasks.

\subsection{Real World Evaluation}

\begin{figure}[t]
    \centering
    \begin{minipage}[t]{0.45\textwidth}
        \centering
        \includegraphics[width=\textwidth]{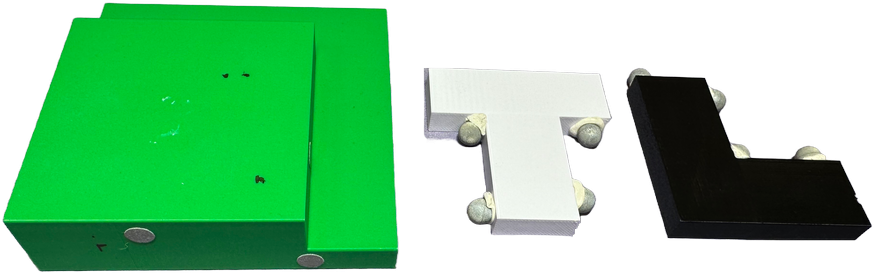}
        \caption{Objects for real-world experiments.}
        \label{fig:objects}
    \end{minipage}
\end{figure}
Finally, we conducted physical experiments using a \panda{} to validate the real-world applicability of our approach. As in the simulation experiment, we evaluated our best method \equationref{eq:full_relaxation}. Our method was evaluated on the three objects in \figref{fig:objects}: a convex square and a non-convex T- and L-shape, with dimensions and weights listed in \tabref{objects}. The objects' poses were tracked using a motion capture system. The robot could use two different pushers.  We set $\epsilon=10^{-4}$ and executed one trajectory per object.

The experimental results demonstrated successful manipulation from the start to the goal pose across all three object geometries. An example execution is presented in \figref{fig:pull_figure}, while the complete execution sequences are available at \url{https://probabilistic-prehensile-pushing.github.io/}. These real-world results confirm that the trajectories generated by our optimization approach are robust enough to transfer from planning to physical execution, even when subject to real-world factors such as friction variations, contact uncertainties, and robot control imperfections.
 
 \begin{table}[tb]
 \caption{Object dimensions and weights.}
 \begin{center}
 \begin{tabular}{lcccc}
 \toprule
  \textbf{Shape} & \multicolumn{2}{c}{\textbf{Dim [L, W, H] (mm)}} & \multicolumn{2}{c}{\textbf{Mass (g)}} \\
 \midrule
 Square  & &100, 20, 100 & &121 \\
T-shape & & 60, 10, 35 & & 47 \\
L-shape & & 60, 10, 60 & & 12\\
 \bottomrule
 \end{tabular}
 \label{objects}
 \end{center}
 \end{table}
\section{Discussion}

Together, our experimental results demonstrate the advantages of our trajectory optimization approach over \baseline{}. The numerical evaluations show that our method finds solutions 8 times faster and achieves costs 20 times lower than \baseline. This improved performance translated directly to practical benefits, with our method achieving a 10x lower goal distance in simulation than RRT-MC. Furthermore, our real-world experiments validated that a physical robot can successfully reorient objects by following the trajectories proposed by our method.

While our method's computation time of approximately 20 seconds precludes real-time feedback control, it offers several advantages. The main benefit is the flexibility to incorporate additional objectives and constraints. For instance, we demonstrated how pusher switches could be minimized by introducing the KL-cost term in \eqref{eq:kl}. Implementing similar behavior modifications in sampling-based methods like \baseline{} would require non-trivial changes to the tree expansion strategy, potentially compromising exploration efficiency and further increasing computation time. Another benefit is our method's reduced sensitivity to hyper-parameters compared to \baseline. We experimentally discovered that for \baseline{}, many step size and $\epsilon$ combinations resulted in overshooting behaviors where a solution was never found. In contrast, our optimization-based approach proved more robust across different parameter settings.

Finally, our probabilistic relaxation of mixed-integer constraints represents a novel theoretical contribution that could be valuable beyond this specific application. For instance, in cluttered pick-and-place scenarios, instead of making binary decisions about which objects to move, a planner could assign probabilities to different objects and smoothly explore various clearing sequences while naturally converging to discrete choices through entropy minimization.

Regarding the computation time, several directions exist for reducing it. One is to design a specialized solver that exploits the problem's structure. Another is to relax the \ac{nlp} in \eqref{eq:relaxation} into a convex problem, which could unlock millisecond solution rates. Finally, warm-starting techniques could be employed for faster re-planning in dynamic scenarios.

Another line of future work is combining \baseline{} and the trajectory optimization formulation. For instance, \baseline{} could generate diverse initial trajectories that our method then refines. This hybrid approach might prove valuable for more complex scenarios like bi-manual prehensile manipulation, where one end-effector is a movable pusher, and the other is a movable support \cite{cruciani2018dexterous}. A final suggestion for future work is to extend the proposed method to closed-loop estimation of friction parameters.
\section{Conclusion}

This paper addressed the challenging task of prehensile pushing. Our solution formulated prehensile pushing as a non-linear trajectory optimization problem using \acp{mc}. The key idea was to relax the binary variables in the original \ac{minlp} into a discrete probability distribution and to penalize the entropy of said distribution, resulting in a quick-to-solve \ac{nlp}. Although the probabilistic formulation allowed the robot to use all pushers simultaneously, the probability mass focused on one distinct pusher at the optimum due to the entropy minimization. The experimental evaluations demonstrated that our trajectory optimization method found solutions 8 times faster and at a 20 times lower cost than a \sota{} sampling-based baseline. Finally, we verified that a simulated and real-world \panda{} robot could successfully reorient objects by following the trajectories proposed by our solution. 
\begin{appendices}
\section{}\label{app:technical_details}
The following parameters simulated prehensile pushing in \mujoco{} \cite{todorov2012mujoco}. The numerical integrator was set to implicitfast, with a stepsize of $0.002$. We enabled the multiccd parameter to allow for multiple-contact collision detection between geom pairs and set condim to 6. The sliding, torsional, and rolling friction were set to 1.05, 0.05, and 0.1, respectively. The solimp parameters $d_0$, $d_{\text{width}}$, width, midpoint, and power were set to $1$, $0.5$, $0.2$, $0.5$, $2$, respectively. The solref parameters timeconst and dampratio were set to $0.001$ and $1$, respectively.
\end{appendices}

\balance
\bibliographystyle{IEEEtran}
\bibliography{references}
\end{document}